\documentclass[11pt,a4paper]{article}
\usepackage[hyperref]{emnlp2020}
\usepackage{times}
\usepackage{latexsym}
\usepackage{graphicx}
\usepackage{subfigure}
\usepackage{multirow}

\usepackage{CJKutf8}
\usepackage{microtype}
\usepackage{amsthm,amsmath,amssymb,bm}
\usepackage{mathrsfs}
\usepackage{algorithm}
\usepackage{algorithmic}

\aclfinalcopy 
\def\aclpaperid{3441} 

\title{Generating Diverse Translation from Model Distribution with Dropout}

\author{Xuanfu Wu$^{1,2}$, \quad Yang Feng$^{1,2}$\thanks{\ \ Corresponding author: Yang Feng},\quad Chenze Shao$^{1,2}$ \\
$^{1}$ Key Laboratory of Intelligent Information Processing\\
Institute of Computing Technology, Chinese Academy of Sciences (ICT/CAS)\\
$^{2}$ University of Chinese Academy of Sciences, Beijing, China\\
{\tt \{wuxuanfu20s, fengyang, shaochenze18z\}@ict.ac.cn }}

\date{}

\begin{document}
\begin{CJK}{UTF8}{gbsn}
\maketitle
\begin{abstract}
Despite the improvement of translation quality, neural machine translation (NMT) often suffers from the lack of diversity in its generation. In this paper, we propose to generate diverse translations by deriving a large number of possible models with Bayesian modelling and sampling models from them for inference. The possible models are obtained by applying concrete dropout to the NMT model and each of them has specific confidence for its prediction, which corresponds to a posterior model distribution under specific training data in the principle of Bayesian modeling. With variational inference, the posterior model distribution can be approximated with a variational distribution, from which the final models for inference are sampled. We conducted experiments on Chinese-English and English-German translation tasks and the results shows that our method makes a better trade-off between diversity and accuracy.

\end{abstract}

\section{Introduction}

In the past several years, neural machine translation (NMT) ~\citep{kalchbrenner2013recurrent, sutskever2014sequence,  gehring-etal-2017-convolutional, vaswani2017attention, zhang2019bridging} based on the end-to-end model has achieved impressive progress in improving the accuracy of translation. Despite its remarkable success, NMT still faces problems in diversity. In natural language, due to lexical, syntactic and synonymous factors, there are usually multiple proper translations for a sentence. However, existing NMT models mostly implement one-to-one mapping between natural languages, that is, one source language sentence corresponds to one target language sentence. Although beam search, a widely used decoding algorithm, can generate a group of translations, its search space is too narrow to extract diverse translations.

There are some researches working at enhancing translation diversity in recent years.  ~\citet{li2016simple} and \citet{vijayakumar2016diverse} proposed to add regularization terms to the beam search algorithm so that it can possess greater diversity.
~\citet{he2018sequence} and \citet{shen2019mixture} introduced latent variables into the NMT model, thus the model can generate diverse outputs using different latent variables. Moreover, ~\citet{sun2019generating} proposed to combine the structural characteristics of Transformer and use the different weights between each head in the multi-head attention mechanism to obtain diverse results. In spite of improvement in balancing accuracy and diversity, these methods do not represent the diversity in the NMT model directly.

In this paper, we take a different approach to generate diverse translation by explicitly maintaining different models based on the principle of Bayesian Neural Networks (BNNs). These models are derived by applying concrete dropout ~\cite{gal2017concrete} to the original NMT model and each of them is given a probability to show its confidence in generation. According to Bayesian theorem, the probabilities over all the possible models under a specific training dataset forms a posterior model distribution which should be involved at inference. To make the posterior model distribution obtainable at inference, we further employ variational inference ~\citep{hinton1993keeping, neal1995bayesian, graves2011practical, blundell2015weight} to infer a variational distribution to approximate it, then at inference we can sample a specific model based on the variational distribution for generation.



We conducted experiments on the NIST Chinese-English and the WMT'14 English-German translation tasks and compared our method with different strong baseline approaches. The experiment results show that our method can get a good trade-off in translation diversity and accuracy with little training cost. 

Our contributions in this paper are as follows:
\begin{itemize}
\item We introduce Bayesian neural networks with variational inference to NMT tasks to explicitly maintain different models for diverse generation. 
\item We apply concrete dropout to the NMT model to derive the possible models which only demands a small cost in computation. 
\end{itemize}



\section{Background}


Assume a source sentence with $n$ words $\bm{x}=x_{1},x_{2},...,x_{n}$, and its corresponding target sentence with $m$ words $\bm{y}=y_{1},y_{2},...,y_{m}$, NMT models the probability of generating $\bm{y}$ with $\bm{x}$ as the input. Based on the encoder-decoder framework, NMT model $\Theta$ encodes source sentence into hidden states by its encoder, and uses its decoder to find the probability of $t$-th word $\bm{y}_{t}$, which depends on the hidden states and the first $t-1$ words of target sentence $\bm{y}$. The translation probability from sentence $\bm{x}$ and $\bm{y}$ can be expressed as:
\begin{equation}\label{equ:prob}
    {P}(\bm{y}|\bm{x}) = 
    \mathop{\large\prod}\limits_{t=1}^m {P}({y}_{t}|{y}_{<t},\bm{x};\Theta)
\end{equation}

Given a training dataset with source-target sentence pairs $ \mathcal{D} = \left\{{(\bm{x_{1}},\bm{y_{1}^*}),...,(\bm{x_{D}},\bm{y_{D}^*})}\right\} $, the loss function we want to minimize in the training is the sum of negative log-likelihood of Equation \ref{equ:prob}:
\begin{equation}\label{equ:loss}
    L = 
    -\mathop{\large\sum}\limits_{(\bm{x_{i}},\bm{y_{i}^*})\in\mathcal{D}} \log{P}(\bm{{y}}=\bm{y_{i}^*}|\bm{{x}}=\bm{x_{i}};\Theta)
\end{equation}

In practice, by properly designing neural network structures and training strategies, we can get the specific model parameters that minimize Equation \ref{equ:loss} and obtain translation results through the model with beam search.

One of the most popular model in NMT is Transformer, which was proposed by ~\citet{vaswani2017attention}. Without recurrent and convolutional networks, Transformer constructs its encoder and decoder by stacking self-attention and fully-connected network layers. 
Self attention is operated with three inputs: query(Q), key(K) and value(V) as:
\begin{equation}\label{equ:attn_base}
    \begin{aligned}
        {\rm Attention}(Q, K, V) = {\rm softmax}(\frac{QK^T}{\sqrt{d_{k}}}V)
    \end{aligned}
\end{equation}
where the dimension of key is $d_{k}$.

Note that Transformer implements the multi-head attention mechanism, projecting inputs into $h$ group inputs to generate $h$ different outputs in Equation \ref{equ:attn_base}, and these outputs are concatenated and projected into final outputs:
\begin{eqnarray}
    head_{i} = {\rm Attention}(QW_{i}^Q, KW_{i}^K, VW_{i}^V) \label{equ:head} \\ 
    {\rm Output} = {\rm Concat}(head_{1},...,head_{h})W^{O} \label{equ:att}
\end{eqnarray}
where $W_{i}^Q$, $W_{i}^K$, $W_{i}^V$ and $W^{O}$ are the projection matrices. The output of Equation \ref{equ:att} is then fed into the fully-connected layer named feed-forward network. The feed-forward network uses two linear networks and a ReLU activation function:
\begin{equation}\label{equ:ff}
    {\rm FFN}(x) = {\rm ReLU}(xW_{1}+b_{1})W_{2}+b_{2}
\end{equation}

We only give a brief description of Transformer above. Please refer to ~\citet{vaswani2017attention} for more details.

\section{Uncertainty Modeling}
\subsection{Bayesian Neural Networks with Variational Inference}

For most of machine learning tasks based on neural networks, a model with specific parameters is trained to explain the observed training data. However, there are usually a large number of possible models that can fit the training data well, which leads to model uncertainty. Model uncertainty may result from noisy data, uncertainty in model parameters or structure uncertainty, and it is represented as the confidence which model to choose to predict with.
In order to express model uncertainty, we consider all possible models with parameters $\bm \omega$ and define a prior distribution $P(\bm \omega)$ over the model (i.e., the space of the parameters) to denote model uncertainty. Then given a training data set $(\bm{X},\bm{Y})$, the predicted distribution for $\bm Y$ can be denoted as $P(\bm{Y}|\bm{X},\bm \omega)$.

Following \citet{gal2017concrete}, we employ Bayesian neural networks (BNNs) to represent the $P(\bm\omega|\bm{X},\bm{Y})$, which is the posterior distribution of the models under $(\bm{X},\bm{Y})$. BNNs offer a probabilistic interpretation of deep learning models by inferring distributions over the models' parameters which are trained using Bayesian inference.
The posterior distribution can be got by invoking Bayes' theorem as:
\begin{equation}\label{equ:bayes_prob}
    \begin{aligned}
        P({\bm \omega}|\bm{X},\bm{Y}) &= \frac{P(\bm{Y}|\bm{X},\bm \omega)P(\bm \omega)}{P(\bm{Y}|\bm{X})}\\ 
        &= \frac{P(\bm{Y}|\bm{X},\bm \omega)P(\bm \omega)}{\mathbb{E}_{\bm \omega}[P(\bm{Y}|\bm{X},\bm \omega)]} .
    \end{aligned}
\end{equation}
Then according to BNNs, given a new test data $\bm{x'}$, the predictive distribution of the output $\bm{y'}$ is:
\begin{equation}\label{equ:bayes_p_prob}
    P(\bm{y'}|\bm{x'},\bm{X},\bm{Y}) = \mathbb{E}_{\bm\omega\sim P(\bm\omega|\bm{X},\bm{Y})}[P(\bm{y'}|\bm{x'},\bm\omega)]
\end{equation}

The expectations in Equation \ref{equ:bayes_prob} and \ref{equ:bayes_p_prob} are integrated over model distribution $\bm\omega$, and the huge space of $\bm\omega$ makes it intractable to obtain the results. Therefore, inspired by ~\citet{hinton1993keeping}, \citet{graves2011practical} proposes a variational approximation method, using a variational distribution $Q(\bm\omega|\theta)$ with the parameters $\theta$ to approximate the posterior distribution $P(\bm\omega|\bm{X},\bm{Y})$. To this end, the training objective is to minimize the Kullback-Leibler (KL) divergence between the model distribution and the posterior distribution $\mathrm{KL}(Q(\bm\omega|\theta)||P({\bm\omega}|\bm{X},\bm{Y}))$. With variational inference, the objective is equivalent to maximizing its evidence lower bound (ELBO), so we get
\begin{equation}\label{equ:elbo}
    \begin{aligned}
        {\theta^*} = \mathop{\arg\max}\limits_{\theta} \   
        &{\mathbb{E}_{Q(\bm\omega|\theta)}\log P(\bm{Y}|\bm{X}, \bm\omega)} \\ &{-\mathrm{KL}((Q(\bm\omega|\theta)||P({\bm\omega}))}\\
    \end{aligned}
\end{equation}

As we can see in Equation \ref{equ:elbo}, the first term on the right side is the expectation of the predicted probability over model distribution on the training set, which can be unbiased estimated with the Monte-Carlo method. And the second term is the KL divergence between the approximate model distribution and the prior distribution. From the perspective of ~\citet{hinton1993keeping} and \citet{graves2011practical}, with the above objective, we can express model uncertainty under the training data and meanwhile regularize model parameters and avoid over-fitting. Therefore, at inference, we can use the distribution $Q(\bm\omega|\theta)$ instead of $P(\bm\omega|\bm{X},\bm{Y})$ to evaluate model confidence (i.e., model uncertainty).

\subsection{Model distribution with Dropout}
%

To derive the BNN, we need to first decide how to explore for the possible models and then decide the prior distribution and the variational distribution for the models. As in \citealp{gal2017concrete}, we can define a simple model with parameters ${\bm\omega_{_{\bm{W}}} (\bm{W} \in \mathbb{R}_{m\times n})}$  and then drop out some column of ${\bm\omega_{_{\bm{W}}}}$ to get the possible models. We use matrix Gaussian distribution as the prior model distribution and Bernoulli distribution as the posterior model distribution. 

Using ${\bm{W}{.j}}$ to denote the $j$-th column of ${\bm{W}}$,
we draw a
matrix Gaussian distribution as the probability distribution of dropping out the $j$-th column as
\begin{equation}\label{equ:prior}
    P(\bm\omega_{_{\bm{W}{.j}}}) \sim \mathcal{MN}(\bm\omega_{_{\bm{W}{.j}}};0,I/l,I/l)
\end{equation}
where $l$ is the hyper-parameter. 

The above matrix Gaussian distribution is used as the prior distribution of the models got by dropping out the $j$-th column.
Then we introduce $\bm{p} \ (\bm{p} \in \mathbb{R}_{1\times n})$ as the probability vector of dropping out the columns of ${\bm\omega_{_{\bm{W}}}}$, which means dropping out the $j$-th column with the probability of $\bm{p}_{j}$, and keeping the $j$-th column unchanged with the probability of $1-\bm{p}_{j}$.
Therefore the posterior model distribution of dropping out the $j$-th column is defined as
\begin{equation}\label{equ:model_dist}
Q(\bm\omega_{_{\bm{W}_{.j}}}|\theta)=\left\{
\begin{array}{rcl}
1-\bm{p}_{j}       &      & {\bm\omega_{_{\bm{W}_{.j}}}=\bm{W}_{.j}}\\
\bm{p}_{j}     &      & {\bm\omega_{_{\bm{W}_{.j}}}=\bm{0}}\\
\end{array} \right. 
\end{equation}
where $\bm{W} \in \theta$ and $\bm{p} \in \theta$ are trainable parameters.

With Equation \ref{equ:prior} as prior, the KL divergence for the $j$-th column of the matrix can be represented as:
\begin{equation}\label{equ:kl_column}
    \begin{aligned}
    &\mathrm{KL}(Q(\omega_{\bm{W}_{.j}}|\theta)||P(\omega_{\bm{W}_{.j}}))\\
    =&\mathcal{R}(\bm{p}_{\bm{W}_{.j}},\bm{W}_{.j},l)-\mathcal{H}(\bm{p}_{\bm{W}_{.j}})\\
    \end{aligned}
\end{equation}
where 
\begin{equation}\label{equ:kl_r}
    \mathcal{R}(\bm{p}_{\bm{W}_{.j}},\bm{W}_{.j},l)=\frac{(1-\bm{p}_{j})l^2}{2}\mathop{\sum}\limits_{i=1}^m\bm{W}_{ij}^2
\end{equation}
and
\begin{equation}\label{equ:kl_h}
    \mathcal{H}(\bm{p}_{\bm{W}_{.j}})=-[\bm{p}_{j}\log(\bm{p}_{j})+(1-\bm{p}_{j})\log(1-\bm{p}_{j})]
\end{equation}

Since the probability distribution among different neural networks and different columns of neural network are independent. For a complex multi-layer neural network $\theta$, the KL divergence between model distribution $Q(\omega|\theta)$ and prior distribution $P({\omega})$ is
\begin{equation}\label{equ:kl_all}
    \begin{aligned}
        &\mathrm{KL}(Q(\omega|\theta)||P({\omega}))\\
        =&\mathop{\large\sum}\limits_{\bm{W}_{m\times n},  \bm{p} \in\theta}\mathop{\large\sum}\limits_{j=1}^n\mathcal{R}(\bm{p}_{\bm{W}_{.j}},\bm{W}_{.j},l)-\mathcal{H}(\bm{p}_{\bm{W}_{.j}})\\
    \end{aligned}
\end{equation}

\section{Application to Transformer}
Previous sections show how to use concrete dropout to realize variational approximation of the posterior model distribution. In this section we will introduce the implementation in representing model distribution for Transformer with aforementioned methods.

\subsection{Dropout in Transformer}

Stated in detail in ~\citet{vaswani2017attention}, in Transformer, dropout is commonly used to the output of modules, including the output of embedding, attention layer and feed-forward layer. Also, from Equation $\ref{equ:kl_r}$, we find it's important to find the network $W$ corresponding to the dropout module. The correspondences in Transformer are as follows:

\textbf{Embedding module} Embedding module works for mapping the words into embedding vectors. The embedding module contains a matrix $W_{E}\in \mathbb{R}_{l_{d}\times d}$, where $l_{d}$ is the length of dictionary and $d$ is the dimension of embedding vector. For the $i$-th word in the dictionary, its embedding vector is the $i$-th column of $W_{E}$. Since dropout the $j$-th dimension of word embedding is equivalent to dropping out the $j$-th row of $W_{E}$, we utilize $W_{E}^T$ and its corresponding dropout in Equation $\ref{equ:kl_r}$. 

\textbf{Attention module} For attention modules, as we can see in $\ref{equ:att}$, their outputs are generated by concatenating the output of different heads and projecting by matrix $W^{O}$. Since dropout is used in the output of attention module, we take $W^{O}$ and its corresponding dropout in calculating Equation $\ref{equ:kl_r}$. 

\textbf{Feed-forward module} As shown in Equation $\ref{equ:ff}$, the output is generated through $W_{2}$ with bias $b_{2}$. As we can see, for network $y=xW+b$, we can find that

\begin{equation}\label{equ:feed-fwd}
    \begin{aligned}
        y=\mathrm{Concat}(x, 1) \cdot \mathrm{Concat}(W^T, b^T)^T
    \end{aligned}
\end{equation}

as we can see, dropout to the output of the feed-forward module can be regraded as dropping out $W_{2}$ and $b_{2}$. So, during training, we use $\mathrm{Concat}(W^T, b^T)^T$ to calculate Equation $\ref{equ:kl_r}$.

\subsection{Training and Inference}

Although dropout is frequently utilized in Transformer, there are some networks in Transformer like $W_{i}^Q$, $W_{i}^K$, $W_{i}^V$ in Equation \ref{equ:head}, and their output is not masked by dropout. So, in our implementation, we obtain the model distribution by fine-tuning the pre-trained model, freezing their parameters and only updating dropout probabilities. Moreover, we choose different trained modules to train their output dropout probability, and in calculating Equation \ref{equ:kl_all}, we only take those trained dropout probabilities into consideration. By allowing dropout probabilities to change, our method can better represent model distributions under the training dataset than the fixed dropout probabilities. The mini-batch training algorithm is expressed in Equation $\ref{alg1}$. It is worth to mention that since we train the model distribution with batches of data, we scale the KL divergence with the proposition of the batch in the entire training dataset.

\begin{algorithm} 
\caption{Mini-batch Training of Bayesian NN using Variational Inference with Dropout in NMT} 
\label{alg1} 
\begin{algorithmic}[1] 
\REQUIRE Training dataset $\mathcal{D}=(X, Y)$ with size $N$, pre-trained model parameter $\theta$, learning rate $\eta$, learning epoch $E$ 
\ENSURE model parameter $\theta$
\STATE Initial $\theta$
\STATE Split $\mathcal{D}$ into $(X_{1},Y_{1}),...,(X_{n},Y_{n})$ with size $M_{1},...,M_{n}$
\STATE $i=0$
\WHILE{$i<E$} 
\FOR{$j=1$ to $n$} 
\STATE sample $\omega'$ from $Q(\omega|\theta)$
\STATE $L=\mathop{\sum}\limits_{(x,y)\in (X_{j},Y_{j})}\mathop{\sum}\limits_{k} \log{P}(y_{k}|y_{<k},x;\omega')$
\STATE $L=L+\frac{M_{j}}{N}[\mathop{\large\sum}\limits_{W_{m \times n}\in\theta}\mathop{\large\sum}\limits_{j=1}^n-\mathcal{H}(p_{_{W_{.j}}})+\mathcal{R}(p_{_{W_{.j}}},W_{.j},l)]$
\STATE $\theta \leftarrow \theta + \eta \frac{\partial}{\partial \theta} L$
\ENDFOR 
\ENDWHILE 

\end{algorithmic} 
\end{algorithm}

During updating the dropout probability, due to the discrete characteristics of the Bernoulli distribution, we cannot directly calculate the gradient of the first term in Equation \ref{equ:elbo} to the dropout probability. So, we adopt concrete dropout, which is used in ~\citet{gal2017concrete}. As a continuous relaxation of dropout, for its input $\bm{y}$, 
 the output can be expressed as $\bm{y'}=\bm{y} \odot \bm{z}$, and vector $\bm{z}$ satisfies:
\begin{equation}\label{equ:conc_drop}
    \begin{aligned}
        &\bm{z}={\rm sigmoid}(\frac{1}{t}({\rm log}(p)-{\rm log}(1-p)+{\rm log}(u)\\
        &-{\rm log}(1-u)))
    \end{aligned}
\end{equation}
where $u \sim \mathcal{U}(0,1)$, $p$ is dropout probability.

In the inference stage, we just randomly mask model parameters with trained dropout probabilities, with different random seeds, NMT models with different parameters are sampled. Since diverse translations are demanded, we performed several forward passes through different sampled NMT models, and different translations are generated with different model outputs and beam search. 

\section{Experiment Setup}

\textbf{Dataset} In our experiment, we select datasets in the following translation tasks:

\hangafter=1
\hangindent 2em
• NIST Chinese-to-English (NIST Zh-En). Its dataset is based on LDC news corpus and contains about 1.34 million sentence pairs. It also includes 6 relatively small datasets, MT02, MT03, MT04, MT05, MT06, and MT08. In our experiments, we use MT02 as the development set, and the rest work as the test sets. Without special explanation, we use average result of test sets as final results. 

\hangafter=1
\hangindent 2em
• WMT'14 English-to-German (WMT'14 En-De). Its dataset comes from the WMT'14 news translation task, which contains about 4.5 million sentence pairs. In our experiment, we use newstest2013 as the development set and newstest2014 as the test set. 

For above two datasets, We adopt Moses tokenizer ~\citep{koehn-etal-2007-moses} in English and German corpus. We also use the byte pair encoding (BPE) algorithm ~\citep{sennrich2015neural}, and limit the size of the vocabulary $K = 32000$. And we train a joint dictionary for WMT'14 En-De. For NIST, we use THULAC toolkit ~\citep{sun2016thulac} to segment Chinese sentence into words. In addition, we remove the examples in datasets from the above two tasks where length of the source language sentence or target language sentence exceed 100 words.  

\textbf{Model Architecture} In our experiments, we all adopt the Transformer Base model in ~\citet{vaswani2017attention}. Transformer base model has 6 layers in encoder and decoder, and it has hidden units with 512 dimension, except for the feed-forward network, where the inner-layer output dimension is 2048. The number of heads in Transformer base model is 8 and the default dropout probability is 0.1. And our model is implemented in python3 with the Fairseq-py ~\citep{ott2019fairseq} toolkit.

\textbf{Experimental Setting} During training, in order to improve the accuracy, we use the label smoothing ~\citep{szegedy2016rethinking} with $\epsilon=0.1$. In terms of optimizer, we adopt the Adam optimizer ~\citep{kingma2014adam}, the main parameters of the optimizer is $\beta_{1} = 0.9$, $\beta_{2} = 0.98$, and $\epsilon = 10^{-9}$. As for the learning rate, we adopt the dynamic learning rate method in \citet{vaswani2017attention} with $\mathrm{warmup\underline{~~}steps}=4000$. Also, we use mini-batch training with $\mathrm{max\underline{~~}token}=4096$.

\textbf{Metrics} In terms of evaluation metrics, referring to ~\citet{shen2019mixture}, we adopt the BLEU and Pairwise-BLEU to evaluate translation quality and diversity. Both two metrics are calculated with case-insensitive BLEU algorithm in ~\citet{papineni-etal-2002-bleu}. In our experiments, the BLEU is to measure the average similarity between the output translations and the standard translation. The higher the BLEU value, the better the accuracy of translation. And the Pairwise-BLEU reflects the average similarity between the output translations of different groups. The lower the Pairwise-BLEU value, the lower the similarities, and the more diverse the translations. In our experiment, we use the NLTK toolkit to calculate the two metrics.

\section{Experiment Results}

\subsection{Analysis of Training Modules and Hyper-parameter}

In this experiment, we train models with different training modules and hyper-parameter $l$ with NIST dataset to evaluate their effects, and some results are shown in Table \ref{citation-guide1}.

\begin{table*}[!t]
\centering
\begin{tabular}{@{}cc|cccccccc@{}}
\hline
\multicolumn{2}{c|}{$l^{2}$}  & $10^{1}$    & $10^{2}$   & $10^{3}$  & $10^{4}$ & $10^{5}$ & $10^{6}$ & $10^{7}$ & $10^{8}$ \\
\hline
\multirow{2}{*}{Decoder's 1-3 Layer} & BLEU          & 42.48 & 42.47 & 42.56 & 41.33 & 26.83  & 8.68    & 5.14     & 2.90  \\
                                   & Pairwise-BLEU & 80.97 & 79.66 & 72.12 & 62.16 & 43.11  & 33.24   & 25.43    & 35.18  \\
\multirow{2}{*}{Decoder}           & BLEU          & 42.34 & 42.30 & 42.25 & 38.17 & 7.79   & 12.22   & 0.59 & 0.26 \\
                                   & Pairwise-BLEU & 75.89 & 74.52 & 67.26 & 49.67 & 16.86 & 20.47 & 56.99 & 58.92 \\
\multirow{2}{*}{Encoder+Decoder}   & BLEU          & 42.20 & 42.12 & 41.69 & 32.70 & 2.43   & 0.83    & 0.37 &  0.68\\
                                   & Pairwise-BLEU & 69.75 & 68.08 & 58.99 & 38.02 & 9.85   & 79.80   & 56.76 & 58.54 \\    

\hline
\end{tabular}
\caption{\label{citation-guide1}
Results with 3 different training modules and hyper-parameter $l$. From the table we can see the BLEU and Pairwise-BLEU change with training modules. Also, the BLEU decreases with the increase of $l$, and the Pairwise-BLEU decreases steadily and then increases when the BLEU value is close to its minimum.
}
\end{table*}

As we can see in Table \ref{citation-guide1}, for those training modules, when $l$ is small, with the same hyper-parameter $l$, choosing smaller part of training modules will lead to lower BLEU and Pairwise-BLEU, showing that accuracy of the generate translations increases while diversity decreases. We also find that in the same training modules, with the increase of $l$, the Pairwise-BLEU decreases steadliy, and then increases when the BLEU is close to zero; and the BLEU has similar trends with Pairwise-BLEU, however, when $l$ is relatively low, the BLEU tends to stablize. 

For the above-mentioned experimental results, we can interpret as follows: in training modules, since Equation $\ref{equ:kl_all}$ is the sum of the training modules' KL divergence, with the training modules increase, the KL divergence accordingly increases, pushing the dropout probability higher and making translations diverse. In terms of hyper-parameter $l$, as we can see in Equation \ref{equ:prior}, when $l$ increases, the prior distribution is squeezed to zero matrix; thus during training, the dropout probabilities will get higher to make the model distribution close to prior distribution. However, when the $l$ is too high to make most of dropout probabilities close to 1, uncertainty of model parameters decreases, making Pairwise-BLEU increases.

\subsection{Results in Diverse Translation}

From the previous section, we can see that by selecting different training modules and hyper-parameter, translations with different accuracy and diversity can be obtained. Then we conduct experiments to generate 5 groups of different translations on NIST Zh-En dataset and WMT'14 En-De dataset, and compare the diversity and accuracy of the translations generated by our method and the following baseline approaches:

\hangafter=1
\hangindent 2em
• Beam Search: we choose the optimal 5 results directly generated by beam search in this paper.

\hangafter=1
\hangindent 2em
• Diverse Beam Search (DBS) ~\citep{vijayakumar2016diverse}: it works by grouping the outputs and adding regularization terms in beam search to encourage diversity. In our experiment, the number of output translations of groups are all 5.

\hangafter=1
\hangindent 2em
• HardMoE~\citep{shen2019mixture}: it trains model with different hidden states and obtains different translations by controlling hidden state. In our experiment, we set the number of hidden states is 5.

\hangafter=1
\hangindent 2em
• Head Sampling~\citep{sun2019generating}: it generate different translations by sampling different encoder-decoder attention heads according to their attention weight, and copying the samples to other heads in some conditions. Here, we set the parameter $\mathrm{K}=3$.

\begin{figure}[p]
\begin{minipage}[htb]{0.9\linewidth}
\centering
\includegraphics[width=6.5cm, height=5.85cm]{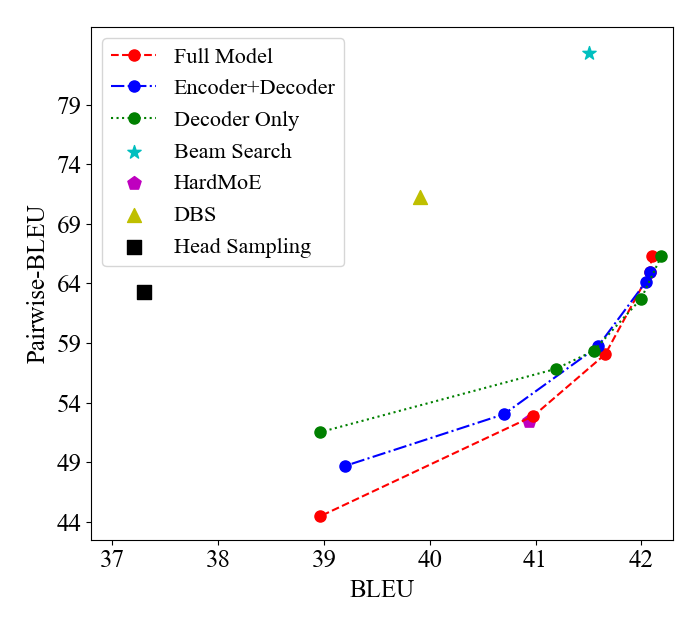}
\end{minipage}
\begin{minipage}[htb]{0.9\linewidth}
\centering
\hspace{2mm}
\includegraphics[width=6.5cm, height=5.85cm]{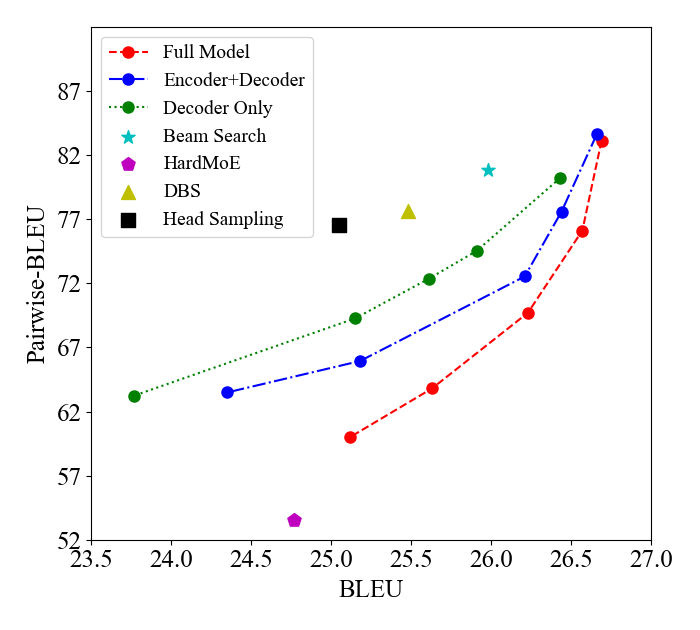}
\end{minipage}
\caption{\label{baseline}
Experiment result in NIST Zh-En (upper one) and WMT'14 En-De (lower one). The X axis and Y axis represents BLEU and Pairwise-BLEU value. The first three groups in legend (connected with curves) are results with our methods under specific training modules with different prior parameters $l$, and the latter four groups (scattered points) are results from baseline methods.}
\end{figure}

\begin{table*}[!h]
\centering
\begin{tabular}{@{}ll@{}}
\hline
Source    & 此次会议的一个重要议题是跨大西洋关系。 \\
\hline
Reference & One of the important topics for discussion in this meeting is the cross atlantic relation. \\
          & One of the top agendas of the meeting is to discuss the cross-atlantic relations.          \\
          & An important item on the agenda of the meeting is the trans-atlantic relations.            \\
          & One of the major topics for the conference this time is transatlantic relations.           \\
Beam search & An important item on the agenda of this meeting is transatlantic relations.                \\
          & An important topic of this conference is transatlantic relations.                          \\
          & An important topic of this meeting is transatlantic relations.                             \\
          & An important topic of this conference is transatlantic ties.                               \\
          & An important topic of the meeting is transatlantic relations.                              \\
Our Work  & One of the important topics of this conference is transatlantic relations.                 \\
          & An important item on the agenda of this meeting is the transatlantic relationship.         \\
          & An important item on the agenda of this conference is the transatlantic relationship.      \\
          & An important topic for discussion at this conference is cross-atlantic relations.          \\
          & One of the important topics of this conference is the transatlantic relationship.         \\
\hline
\end{tabular}
\caption{\label{citation-guide2}
Translation examples in NIST Zh-En Task. Results of our work is generated by training dropout in decoder with $l^2=1000$. The result shows that by adjusting model parameters, our method can generate translations with higher diversity while maintaining accuracy.
}
\end{table*}

The results are shown in Figure \ref{baseline}. In Figure \ref{baseline} we plot the BLEU versus Pairwise-BLEU, the scattered points show the results of baseline approaches, and the points on the curves are results in the same training modules with different hyper-parameter $l$. From Figure \ref{baseline}, firstly, we can verity that choosing different training modules can lead to different balance of translation diversity and accuracy, for NIST Zh-En and WMT'14 En-De, training dropout probabilities in the full model can get better translations.

Also, we suggest that in NIST Zh-En task, by adjusting training modules and hyper-parameter $l$, our results which has higher BLEU and lower Pairwise-BLEU values than baselines without HardMoE, even for HardMoE itself, our method is comparable with proper $l$ while training the whole model. In WMT'14 En-De, we also find that our method exceeds the baseline approach except HardMoE. For the gap in performace with HardMoE, we interpreted that since our models are randomly sampled from the model distribution, it could be hard for our models to represent such distinguishable characteristics like HardMoE, which trains multiple different latent variables.

Also, to intuitively display the improvement of diversity in our translations, we choose a case from NIST Zh-En task, the results are shown in Table \ref{citation-guide2}. The case shows that compared with beam search, which only varies in few words, our method can obtain more diverse translations while ensuring the accuracy of translation, and diversities are not only shown in words, but also reflected in lexical characteristic.

\subsection{Analyzing Module Importance with Dropout Probability}

Some researches ~\citep{voita2019analyzing, michel2019sixteen, fan2019reducing} found that a well-trained Transformer model is over-parameterized. Useful information gathers in some parameters and some modules and layers can be pruned to improve the efficiency during test time. Since dropout can play the role of regularization and there are differences in the trained dropout probabilities of different neuron, we conjecture that the trained dropout probability and the importance of each module are correlated.
To investigate this, we choose the model in which dropout probabilities of the full model is trained with $l^2=400$ in NIST Zh-En task, and separately calculate the average dropout probability $\bar{p}_{dropout}$ of different attention modules. Also, we manually pruned the corresponding modules of the model, obtained translations and calculated its BLEU. the more the BLEU drops, the more important the module is to translation. To quantify their relevance, we calculate the Pearson correlation coefficient (PCC) $\rho$ in different kinds of training modules, and highlights the highest and lowest results.

\begin{table*}[h]
\centering
\begin{tabular}{@{}c|cc|cc|cc@{}}
\hline
& \multicolumn{2}{c}{Encoder}& \multicolumn{4}{c}{Decoder}\\

& \multicolumn{2}{c}{Self-attention} & \multicolumn{2}{c}{Self-attention} & \multicolumn{2}{c}{E-D Attention} \\
Layer & $\bar{p}_{dropout}$ & BLEU & $\bar{p}_{dropout}$ & BLEU & $\bar{p}_{dropout}$ & BLEU \\
\hline
1 & \textbf{0.0400} & \textbf{32.65} & 0.0484 & 40.20 & \textbf{0.0915} & \textbf{42.15} \\
2 & 0.0858 & 40.97 & \textbf{0.0793} & \textbf{41.67} & 0.0798 & 41.03\\
3 & \textbf{0.0863} & \textbf{41.70} & 0.0670 & 40.83 & 0.0620 & 35.65 \\
4 & 0.0763 & 39.87 & \textbf{0.0460} & 39.56 & 0.0556 & 37.29 \\
5 & 0.0632 & 40.17 & 0.0476 & \textbf{37.93} & 0.0394 & 32.18 \\
6 & 0.0769 & 39.15 & 0.0490 & 40.88 & \textbf{0.0335} & \textbf{18.48} \\
\hline
$\rho$ & \multicolumn{2}{c|}{0.919}         & \multicolumn{2}{c|}{0.689}          & \multicolumn{2}{c}{0.858}   \\
\hline
\end{tabular}
\caption{\label{citation-guide3}
Average dropout probabilities of each module and BLEU of translations generated by model where the module is pruned. From the maximum and minimum of $\bar{p}_{dropout}$ and BLEU, and correlation coefficient $\rho$ in different modules, it is obvious that dropout probabilities of module and its importance is correlated.
}
\end{table*}

Results of our experiment are shown in Table \ref{citation-guide3}. Firstly, we can see that the average dropout probabilities and BLEU are not fully positively correlated, which might be explained by the contingency of sampling from model distribution during training. But from the maximum and minimum of $\bar{p}_{dropout}$ and BLEU, we can find that the dropout probabilities $\bar{p}_{dropout}$ and the BLEU of translations show some similar information in module importance. Also, we quantify the correlation between the $\bar{p}_{dropout}$ and BLEU, finding that it is highly correlated in self-attention module in encoder and in E-D attention in decoder, since its correlation coefficient $\rho>0.8$, and the $\bar{p}_{dropout}$ and BLEU is also correlated in self-attention in decoder. 

\section{Related Work}

Researches in Bayesian Neural Network have a long history, ~\citet{hinton1993keeping} firstly proposes a variational inference approximation methods to BNN to minimize the minimum description length (MDL), then ~\citet{neal1995bayesian} approximate BNN by Hamiltonian Monte Carlo methods. In recent years, ~\citet{graves2011practical} introduces the concept of variational inference, by approximating posterior distribution with model distribution, the model minimizes its MDL and reduces the model weight; and ~\citet{blundell2015weight} proposes an algorithm similar to ~\citet{graves2011practical}, however, it uses mixture of Gaussian densities as prior and achieved comparable performance with dropout in regularization.

Introduced by ~\citet{hinton2012improving}, dropout, which is easy to implement, works as a stochastic regularization to avoid over-fitting. And there are several theoretical explainations such as getting sufficient model combinations ~\citep{hinton2012improving, srivastava2014dropout} to train and augumenting training data ~\citep{bouthillier2015dropout}. ~\citet{gal2016dropout} proposes that dropout can be understood as a bayesian inferences algorithm, and ~\citet{gal2017concrete} uses concrete dropout in updating dropout probabilities. Also, the author implements the dropout methods to represent uncertainty in different kinds of deep learning tasks in ~\citet{gal2016uncertainty}.

In neural machine translation task, lack of diversity is a widely acknowledged problem, some researches like ~\citet{ott2018analyzing} investigate the cause of uncertainty in NMT, and some provide metrics to evaluate the translation uncertainty like ~\citet{galley2015deltableu, dreyer2012hyter}. There are also other researches that put forward methods to obtain diverse translation. ~\citet{li2016simple,vijayakumar2016diverse} adjust decoding algorithms, adding different kinds of diversity regularization terms to encourage generating diverse outputs.
~\citet{he2018sequence,shen2019mixture} utilize mixture of experts (MoE) method, using differentiated latent variables to control generation of translation. ~\citet{sun2019generating} generates diverse translation by sampling heads in encoder-decoder attention module in Transformer model, since different heads may present different target-source alignment. ~\citet{shu2019generating} uses sentence codes to condition translation generation and obtain diverse translations.
\citet{shao2018greedy} propose a new probabilistic ngram-based loss to conduct sequence-level training for generating diverse translation.\citet{feng2020modeling} propose to employ future information to evaluate fluency and faithfulness to encourage diverse translation. 

There are also a few papers in interpreting Transformer model, ~\citet{voita2019analyzing} suggests that some heads play a consistent role in machine translation, and their roles can be interpreted linguistically; also, they implement $L_{0}$ penalty to prune heads. ~\citet{michel2019sixteen} shows that huge amounts of heads in Transformer can be pruned, and the importance of head is cross-domain. Also, ~\citet{fan2019reducing} shows that the layers in Transformer are also able to be pruned: similar to our work, during training, they drop the whole layer with dropout and trained their probability; however, variational inference strategy is not used in their paper, and they take different kinds of inference strategies to balance performance and efficiency rather than sampling.

\section{Conclusion}

In this paper, we propose to utilize variational inference in diverse machine translation tasks. We represent the Transformer model distribution with dropout, and train the model distributions to minimize its distance to the posterior distribution under specific training dataset. Then we generate diverse translations with the models sampled from the trained model distribution. We further analyze the correlations between module importance and trained dropout probabilities. Experiment results in Chinese-English and English-German translation tasks suggest that by properly adjusting trained modules and prior parameters, we can generate translations which balance accuracy and diversity well. 

In future work, firstly, since our model is randomly sampled from model distribution to generate diverse translation, it is meaningful to explore better algorithms and training strategies to represent model distribution and search for the most distinguishable results in model distribution. Also, we'll try to extend our methods in a wider range of NLP tasks. 

\section*{Acknowledgments}

We thank the anonymous reviewers for their insightful comments and suggestions. This paper was supported by National Key R\&D Program of China（NO. 2017YFE0192900).

\bibliography{anthology,emnlp2020}

\begin{thebibliography}{36}
\expandafter\ifx\csname natexlab\endcsname\relax\def\natexlab#1{#1}\fi

\bibitem[{Blundell et~al.(2015)Blundell, Cornebise, Kavukcuoglu, and
  Wierstra}]{blundell2015weight}
Charles Blundell, Julien Cornebise, Koray Kavukcuoglu, and Daan Wierstra. 2015.
\newblock Weight uncertainty in neural networks.
\newblock \emph{arXiv preprint arXiv:1505.05424}.

\bibitem[{Bouthillier et~al.(2015)Bouthillier, Konda, Vincent, and
  Memisevic}]{bouthillier2015dropout}
Xavier Bouthillier, Kishore Konda, Pascal Vincent, and Roland Memisevic. 2015.
\newblock Dropout as data augmentation.
\newblock \emph{arXiv preprint arXiv:1506.08700}.

\bibitem[{Dreyer and Marcu(2012)}]{dreyer2012hyter}
Markus Dreyer and Daniel Marcu. 2012.
\newblock Hyter: Meaning-equivalent semantics for translation evaluation.
\newblock In \emph{Proceedings of the 2012 Conference of the North American
  Chapter of the Association for Computational Linguistics: Human Language
  Technologies}, pages 162--171. Association for Computational Linguistics.

\bibitem[{Fan et~al.(2019)Fan, Grave, and Joulin}]{fan2019reducing}
Angela Fan, Edouard Grave, and Armand Joulin. 2019.
\newblock Reducing transformer depth on demand with structured dropout.
\newblock \emph{arXiv preprint arXiv:1909.11556}.

\bibitem[{Feng et~al.(2020)Feng, Xie, Gu, Shao, Zhang, Yang, and
  Yu}]{feng2020modeling}
Yang Feng, Wanying Xie, Shuhao Gu, Chenze Shao, Wen Zhang, Zhengxin Yang, and
  Dong Yu. 2020.
\newblock Modeling fluency and faithfulness for diverse neural machine
  translation.
\newblock In \emph{Proceedings of the AAAI Conference on Artificial
  Intelligence}, volume~34, pages 59--66.

\bibitem[{Gal(2016)}]{gal2016uncertainty}
Yarin Gal. 2016.
\newblock Uncertainty in deep learning.
\newblock \emph{University of Cambridge}, 1:3.

\bibitem[{Gal and Ghahramani(2016)}]{gal2016dropout}
Yarin Gal and Zoubin Ghahramani. 2016.
\newblock Dropout as a bayesian approximation: Representing model uncertainty
  in deep learning.
\newblock In \emph{international conference on machine learning}, pages
  1050--1059.

\bibitem[{Gal et~al.(2017)Gal, Hron, and Kendall}]{gal2017concrete}
Yarin Gal, Jiri Hron, and Alex Kendall. 2017.
\newblock Concrete dropout.
\newblock In \emph{Advances in neural information processing systems}, pages
  3581--3590.

\bibitem[{Galley et~al.(2015)Galley, Brockett, Sordoni, Ji, Auli, Quirk,
  Mitchell, Gao, and Dolan}]{galley2015deltableu}
Michel Galley, Chris Brockett, Alessandro Sordoni, Yangfeng Ji, Michael Auli,
  Chris Quirk, Margaret Mitchell, Jianfeng Gao, and Bill Dolan. 2015.
\newblock deltableu: A discriminative metric for generation tasks with
  intrinsically diverse targets.
\newblock \emph{arXiv preprint arXiv:1506.06863}.

\bibitem[{Gehring et~al.(2017)Gehring, Auli, Grangier, and
  Dauphin}]{gehring-etal-2017-convolutional}
Jonas Gehring, Michael Auli, David Grangier, and Yann Dauphin. 2017.
\newblock \href {https://doi.org/10.18653/v1/P17-1012} {A convolutional encoder
  model for neural machine translation}.
\newblock In \emph{Proceedings of the 55th Annual Meeting of the Association
  for Computational Linguistics (Volume 1: Long Papers)}, pages 123--135,
  Vancouver, Canada. Association for Computational Linguistics.

\bibitem[{Graves(2011)}]{graves2011practical}
Alex Graves. 2011.
\newblock Practical variational inference for neural networks.
\newblock In \emph{Advances in neural information processing systems}, pages
  2348--2356.

\bibitem[{He et~al.(2018)He, Haffari, and Norouzi}]{he2018sequence}
Xuanli He, Gholamreza Haffari, and Mohammad Norouzi. 2018.
\newblock Sequence to sequence mixture model for diverse machine translation.
\newblock \emph{arXiv preprint arXiv:1810.07391}.

\bibitem[{Hinton et~al.(2012)Hinton, Srivastava, Krizhevsky, Sutskever, and
  Salakhutdinov}]{hinton2012improving}
Geoffrey~E Hinton, Nitish Srivastava, Alex Krizhevsky, Ilya Sutskever, and
  Ruslan~R Salakhutdinov. 2012.
\newblock Improving neural networks by preventing co-adaptation of feature
  detectors.
\newblock \emph{arXiv preprint arXiv:1207.0580}.

\bibitem[{Hinton and Van~Camp(1993)}]{hinton1993keeping}
Geoffrey~E Hinton and Drew Van~Camp. 1993.
\newblock Keeping the neural networks simple by minimizing the description
  length of the weights.
\newblock In \emph{Proceedings of the sixth annual conference on Computational
  learning theory}, pages 5--13.

\bibitem[{Kalchbrenner and Blunsom(2013)}]{kalchbrenner2013recurrent}
Nal Kalchbrenner and Phil Blunsom. 2013.
\newblock Recurrent continuous translation models.
\newblock In \emph{Proceedings of the 2013 Conference on Empirical Methods in
  Natural Language Processing}, pages 1700--1709.

\bibitem[{Kingma and Ba(2014)}]{kingma2014adam}
Diederik~P Kingma and Jimmy Ba. 2014.
\newblock Adam: A method for stochastic optimization.
\newblock \emph{arXiv preprint arXiv:1412.6980}.

\bibitem[{Koehn et~al.(2007)Koehn, Hoang, Birch, Callison-Burch, Federico,
  Bertoldi, Cowan, Shen, Moran, Zens, Dyer, Bojar, Constantin, and
  Herbst}]{koehn-etal-2007-moses}
Philipp Koehn, Hieu Hoang, Alexandra Birch, Chris Callison-Burch, Marcello
  Federico, Nicola Bertoldi, Brooke Cowan, Wade Shen, Christine Moran, Richard
  Zens, Chris Dyer, Ond{\v{r}}ej Bojar, Alexandra Constantin, and Evan Herbst.
  2007.
\newblock \href {https://www.aclweb.org/anthology/P07-2045} {{M}oses: Open
  source toolkit for statistical machine translation}.
\newblock In \emph{Proceedings of the 45th Annual Meeting of the Association
  for Computational Linguistics Companion Volume Proceedings of the Demo and
  Poster Sessions}, pages 177--180, Prague, Czech Republic. Association for
  Computational Linguistics.

\bibitem[{Li et~al.(2016)Li, Monroe, and Jurafsky}]{li2016simple}
Jiwei Li, Will Monroe, and Dan Jurafsky. 2016.
\newblock A simple, fast diverse decoding algorithm for neural generation.
\newblock \emph{arXiv preprint arXiv:1611.08562}.

\bibitem[{Michel et~al.(2019)Michel, Levy, and Neubig}]{michel2019sixteen}
Paul Michel, Omer Levy, and Graham Neubig. 2019.
\newblock Are sixteen heads really better than one?
\newblock In \emph{Advances in Neural Information Processing Systems}, pages
  14014--14024.

\bibitem[{Neal(1995)}]{neal1995bayesian}
Radford~M Neal. 1995.
\newblock \emph{Bayesian learning for neural networks}.
\newblock Ph.D. thesis, University of Toronto.

\bibitem[{Ott et~al.(2018)Ott, Auli, Grangier, and Ranzato}]{ott2018analyzing}
Myle Ott, Michael Auli, David Grangier, and Marc'Aurelio Ranzato. 2018.
\newblock Analyzing uncertainty in neural machine translation.
\newblock \emph{arXiv preprint arXiv:1803.00047}.

\bibitem[{Ott et~al.(2019)Ott, Edunov, Baevski, Fan, Gross, Ng, Grangier, and
  Auli}]{ott2019fairseq}
Myle Ott, Sergey Edunov, Alexei Baevski, Angela Fan, Sam Gross, Nathan Ng,
  David Grangier, and Michael Auli. 2019.
\newblock fairseq: A fast, extensible toolkit for sequence modeling.
\newblock In \emph{Proceedings of NAACL-HLT 2019: Demonstrations}.

\bibitem[{Papineni et~al.(2002)Papineni, Roukos, Ward, and
  Zhu}]{papineni-etal-2002-bleu}
Kishore Papineni, Salim Roukos, Todd Ward, and Wei-Jing Zhu. 2002.
\newblock \href {https://doi.org/10.3115/1073083.1073135} {{B}leu: a method for
  automatic evaluation of machine translation}.
\newblock In \emph{Proceedings of the 40th Annual Meeting of the Association
  for Computational Linguistics}, pages 311--318, Philadelphia, Pennsylvania,
  USA. Association for Computational Linguistics.

\bibitem[{Sennrich et~al.(2015)Sennrich, Haddow, and
  Birch}]{sennrich2015neural}
Rico Sennrich, Barry Haddow, and Alexandra Birch. 2015.
\newblock Neural machine translation of rare words with subword units.
\newblock \emph{arXiv preprint arXiv:1508.07909}.

\bibitem[{Shao et~al.(2018)Shao, Chen, and Feng}]{shao2018greedy}
Chenze Shao, Xilin Chen, and Yang Feng. 2018.
\newblock Greedy search with probabilistic n-gram matching for neural machine
  translation.
\newblock In \emph{Proceedings of the 2018 Conference on Empirical Methods in
  Natural Language Processing}, pages 4778--4784.

\bibitem[{Shen et~al.(2019)Shen, Ott, Auli, and Ranzato}]{shen2019mixture}
Tianxiao Shen, Myle Ott, Michael Auli, and Marc'Aurelio Ranzato. 2019.
\newblock Mixture models for diverse machine translation: Tricks of the trade.
\newblock \emph{arXiv preprint arXiv:1902.07816}.

\bibitem[{Shu et~al.(2019)Shu, Nakayama, and Cho}]{shu2019generating}
Raphael Shu, Hideki Nakayama, and Kyunghyun Cho. 2019.
\newblock Generating diverse translations with sentence codes.
\newblock In \emph{Proceedings of the 57th Annual Meeting of the Association
  for Computational Linguistics}, pages 1823--1827.

\bibitem[{Srivastava et~al.(2014)Srivastava, Hinton, Krizhevsky, Sutskever, and
  Salakhutdinov}]{srivastava2014dropout}
Nitish Srivastava, Geoffrey Hinton, Alex Krizhevsky, Ilya Sutskever, and Ruslan
  Salakhutdinov. 2014.
\newblock Dropout: a simple way to prevent neural networks from overfitting.
\newblock \emph{The journal of machine learning research}, 15(1):1929--1958.

\bibitem[{Sun et~al.(2016)Sun, Chen, Zhang, Guo, and Liu}]{sun2016thulac}
Maosong Sun, Xinxiong Chen, Kaixu Zhang, Zhipeng Guo, and Zhiyuan Liu. 2016.
\newblock Thulac: An efficient lexical analyzer for chinese.

\bibitem[{Sun et~al.(2019)Sun, Huang, Wei, Dai, and Chen}]{sun2019generating}
Zewei Sun, Shujian Huang, Hao-Ran Wei, Xin-yu Dai, and Jiajun Chen. 2019.
\newblock Generating diverse translation by manipulating multi-head attention.
\newblock \emph{arXiv preprint arXiv:1911.09333}.

\bibitem[{Sutskever et~al.(2014)Sutskever, Vinyals, and
  Le}]{sutskever2014sequence}
Ilya Sutskever, Oriol Vinyals, and Quoc~V Le. 2014.
\newblock Sequence to sequence learning with neural networks.
\newblock In \emph{Advances in neural information processing systems}, pages
  3104--3112.

\bibitem[{Szegedy et~al.(2016)Szegedy, Vanhoucke, Ioffe, Shlens, and
  Wojna}]{szegedy2016rethinking}
Christian Szegedy, Vincent Vanhoucke, Sergey Ioffe, Jon Shlens, and Zbigniew
  Wojna. 2016.
\newblock Rethinking the inception architecture for computer vision.
\newblock In \emph{Proceedings of the IEEE conference on computer vision and
  pattern recognition}, pages 2818--2826.

\bibitem[{Vaswani et~al.(2017)Vaswani, Shazeer, Parmar, Uszkoreit, Jones,
  Gomez, Kaiser, and Polosukhin}]{vaswani2017attention}
Ashish Vaswani, Noam Shazeer, Niki Parmar, Jakob Uszkoreit, Llion Jones,
  Aidan~N Gomez, {\L}ukasz Kaiser, and Illia Polosukhin. 2017.
\newblock Attention is all you need.
\newblock In \emph{Advances in neural information processing systems}, pages
  5998--6008.

\bibitem[{Vijayakumar et~al.(2016)Vijayakumar, Cogswell, Selvaraju, Sun, Lee,
  Crandall, and Batra}]{vijayakumar2016diverse}
Ashwin~K Vijayakumar, Michael Cogswell, Ramprasath~R Selvaraju, Qing Sun,
  Stefan Lee, David Crandall, and Dhruv Batra. 2016.
\newblock Diverse beam search: Decoding diverse solutions from neural sequence
  models.
\newblock \emph{arXiv preprint arXiv:1610.02424}.

\bibitem[{Voita et~al.(2019)Voita, Talbot, Moiseev, Sennrich, and
  Titov}]{voita2019analyzing}
Elena Voita, David Talbot, Fedor Moiseev, Rico Sennrich, and Ivan Titov. 2019.
\newblock Analyzing multi-head self-attention: Specialized heads do the heavy
  lifting, the rest can be pruned.
\newblock \emph{arXiv preprint arXiv:1905.09418}.

\bibitem[{Zhang et~al.(2019)Zhang, Feng, Meng, You, and
  Liu}]{zhang2019bridging}
Wen Zhang, Yang Feng, Fandong Meng, Di~You, and Qun Liu. 2019.
\newblock Bridging the gap between training and inference for neural machine
  translation.
\newblock In \emph{Proceedings of the 57th Annual Meeting of the Association
  for Computational Linguistics}, pages 4334--4343.

\end{thebibliography}
\bibliographystyle{acl_natbib}

\end{CJK}
\end{document}